\documentclass{svproc}
\usepackage{url}
\usepackage{graphicx}
\usepackage{biblatex}
\newcommand{\repeatthanks}{\textsuperscript{\thefootnote}}

\begin{document}
\mainmatter              
\title{Hierarchical Neural Network Approaches for Long Document Classification}
\titlerunning{Hierarchical Approaches for Long Document Classification}  
%
\author{Snehal Khandve\inst{1}\thanks{Authors contributed equally} \and 
Vedangi Wagh\inst{1}\repeatthanks \and
Apurva Wani\inst{1}\repeatthanks \and
Isha Joshi\inst{1} \repeatthanks\and
Raviraj Joshi\inst{2}}
\authorrunning{Snehal Khandve et al.} 
%
\tocauthor{Snehal Khandve, Vedangi Wagh, Apurva Wani, Isha Joshi,
and Raviraj Joshi}
\institute{Pune Institute of Computer Technology, India
\and
Indian Institute of Technology Madras, India}

\maketitle              

\begin{abstract}
Text classification algorithms investigate the intricate relationships between words or phrases and attempt to deduce the document's interpretation. In the last few years, these algorithms have progressed tremendously. Transformer architecture and sentence encoders have proven to give superior results on natural language processing tasks. But a major limitation of these architectures is their applicability for text no longer than a few hundred words. In this paper, we explore hierarchical transfer learning approaches for long document classification. We employ pre-trained Universal Sentence Encoder (USE) and Bidirectional Encoder Representations from Transformers (BERT) in a hierarchical setup to capture better representations efficiently. Our proposed models are conceptually simple where we divide the input data into chunks and then pass this through base models of BERT and USE. Then output representation for each chunk is then propagated through a shallow neural network comprising of LSTMs or CNNs for classifying the text data. These extensions are evaluated on 6 benchmark datasets.
We show that USE + CNN/LSTM performs better than its stand-alone baseline. Whereas the BERT + CNN/LSTM performs on par with its stand-alone counterpart. However, the hierarchical BERT models are still desirable as it avoids the quadratic complexity of the attention mechanism in BERT. Along with the hierarchical approaches, this work also provides a comparison of different deep learning algorithms like USE, BERT, HAN, Longformer, and BigBird for long document classification. The Longformer approach consistently performs well on most of the datasets.

\end{abstract}
\section{Introduction}
An enormous amount of data is generated every day in the form of text. This data is generally unstructured and hence it is necessary to classify it to extract meaningful information. Text classification involves labeling this raw input data into various classes based on their content. There are many machine learning models available that automate and simplify the text classification process \cite{minaee2021deep,wagh2021comparative,wani2021evaluating}.

Processing the text in the document and allocating it to one of the predefined categories is called document classification. A major hurdle in document classification is the large length of the input text. It is difficult to comprehend the entire information in large news articles or movie reviews and categorize them. We begin with existing transformer-based models and then move towards hierarchical models which make use of the inherent structure of the document to perform classification seamlessly \cite{zheng2018hierarchical,bora2021icodenet,joshi2021evaluation}. A document can be seen as a sequence of sentences. The sentences are in turn are a sequence of words. We exploit this structure of words -$>$ sentences -$>$ document and process it using hierarchical neural network models. The first level processes the sequence of words to generate sentence representation whereas the second level processes the sequence of sentence representation to generate document representation. Although such models have been previously studied in literature we present a more exhaustive analysis. In this paper, we make use of two sentence encoder architectures - USE and BERT - which are used to generate sentence embeddings in a hierarchical network. These sentence encoders are pre-trained models thus allowing us to leverage transfer learning with hierarchical models.

Universal Sentence Encoder \cite{cer2018universal} encodes sentences into fixed-length embedding vectors for transfer learning tasks. This model is pre-trained on a general corpus and can be efficiently finetuned for diverse transfer tasks. The paper proposes two models - one uses transformers \cite{vaswani2017attention} and the other uses DAN(Deep Averaging Network) \cite{iyyer2015deep}. The DAN model is highly efficient with linear complexity and gives slightly lower accuracy than the transformer model. We make use of the DAN variation for all the experiments. Since it is difficult to encode long documents into fixed-size vectors we explore a two-level architecture for USE.

The transformer architecture has shown superior results on almost all-natural language processing tasks. Transformers have surpassed the previous sequential models in both efficiency and accuracy. They have the advantage of processing all input tokens simultaneously thereby utilizing the hardware most optimally. This architecture is based on the self-attention mechanism \cite{bahdanau2014neural} where each token is processed to generate a contextual embedding by taking every other token. 
BERT \cite{devlin2018bert} is a novel architecture with a set of bidirectional transformers stacked on top of each other. This model achieved state-of-the-art results on various NLP tasks. Although transformer-based architectures like BERT have a decided advantage over sequential models, they are computationally expensive as the input length increases. The quadratic complexity of the self-attention mechanism is a major limitation for a task like long document classification.


In this paper, we introduce 4 models that have been built upon USE and BERT architectures. We split the input text into segments and obtain USE and BERT representations of these segments using a shared encoder. This is followed by a shallow CNN or LSTM network to perform further classification.

The six datasets that were employed in this work are standard text classification datasets viz., BBC News, AG News, 20 NewsGroup (20 NG), R8, BBC Sports, and IMDB.\\

The novel contributions of this work are as follows:
\begin{itemize}
    \item To the best of our knowledge, this is the first work to use USE in a hierarchical setup along with CNN or LSTM and show its effectiveness.
    \item Although hierarchical approaches have been explored for BERT, we provide a more exhaustive analysis on six benchmark datasets and different deep learning algorithms. We also contrast BERT + LSTM with BERT + CNN to show that the former works better in most of the cases.
    \item We benchmark different deep learning algorithms like USE, BERT, HAN, Longformer, and BigBird on these datasets. The Longformer consistently performs well on most of the datasets.
\end{itemize}

\section{Related Work}

In the pre-Transformers era, the deep learning approaches for text classification typically made use of CNNs or RNNs. Text classification using CNNs with little hyperparameter tuning showed significant improvement over traditional approaches \cite{chen2015convolutional}. The performance was further improved with deeper CNNs which were implemented at the character level \cite{conneau2016very,zhang2015character} instead of word level. \cite{socher2013recursive,tai2015improved} make use of tree-structured LSTMs for text classification. Architectures with combinations of LSTM and CNN are used in \cite{lai2015recurrent} \cite{zhou2015c} \cite{xiao2016efficient}. In \cite{tang2015document} a hierarchical network is implemented using simple CNN and RNN for document classification. In the first stage, a sentence is represented using an LSTM or CNN. These sentence representations are passed to gated RNNs to form a vector representing the entire document.

Hierarchical Attention Networks(HAN) were introduced in \cite{yang2016hierarchical} which used self-attention \cite{bahdanau2014neural} at the word-level and sentence-level. The model mirrored the inherent structure of documents in the form of paragraphs, sentences, and words. \cite{gao2018hierarchical} implemented Hierarchical Convolutional Attention Networks (HCAN) which are like transformers \cite{vaswani2017attention}, used convoluted self-attention layers on raw embeddings. This approach could be trained faster than RNNs and gave competitive results. One of the shortcomings of transformers is that they do not perform well when it comes to classifying long sequences of text. It was addressed by introducing two models viz. Tobert(Transformer over BERT) and RoBERT(Recurrence over BERT),  built on the top of BERT \cite{devlin2018bert}. The input text was divided into smaller chunks, passed on to the base model of BERT to obtain their word representations which are then given to the next layer(LSTM or BERT)\cite{pappagari2019hierarchical}. \cite{barnes2020hierarchical} performed an empirical comparison of hierarchical models and transfer learning methods for document classification in five languages. Hierarchical networks are seen to outperform previous baselines and this effect is particularly stronger for longer documents.

CNNs have also been explored extensively for text classification. Recording CNN results as baseline accuracies, \cite{ive2018hierarchical} used RNNs to take into account the sequential nature of events occurring in the mental health-related text. They also studied attention mechanisms to focus only on the relevant text. Thus Hierarchical Recurrent Neural Network (RNN) architecture was built to classify the text with improved performance. This idea is further extended by \cite{chang2019language}, by proposing language-model-based pre-training methods for hierarchical document representations. The two methods proposed are: 1) Pre-training hierarchical left-to-right and right-to-left document representations and, 2) Masked language model technique to pre-train bidirectional hierarchical document-level representations.

BERT \cite{devlin2018bert} is a set of bidirectional stacked transformer \cite{vaswani2017attention} units that are trained to perform two tasks namely masked word prediction and next sentence prediction. BERT has been used for document classification \cite{adhikari2019docbert} but the input length was truncated to 512, the typical input length for BERT. To overcome the limitation of quadratic complexity posed by the self-attention mechanism, some variations of BERT have been implemented to classify longer documents without increasing the complexity \cite{kitaev2020reformer}. \cite{beltagy2020longformer,zaheer2020big} use simultaneous global attention and local sliding windows thereby reducing complexity. \cite{dai2019transformer} connects multiple segments of text by recurrence and uses relative position embeddings.

\section{Dataset Details}
We make use of text classification datasets with longer document lengths \cite{wagh2021comparative}. The datasets used in this work are described below.
\begin{itemize}
    \item \textbf{20NG} :
The 20 Newsgroup dataset \cite{scikit-learn} contains a total of 18774 records. Using an 80\% split, train data contains 11314 records and test data contains 7532 records. Each record is a news article with an average size of 315 words. There are 20 distinct classes in this dataset.
\item \textbf{BBC News} :
\cite{greene06icml} dataset is one of the popular text classification datasets. It contains news articles classified into 5 different categories viz., Sport, Business, Politics, Tech, and Entertainment. There are 2225 records each with an average size of 389 words.
\item \textbf{AG News} :
\cite{AGNews:online} dataset is divided into a training set with 120000 records and a test set with 7600 records. Each record is a news article collected by ComeToMyHead. There are 4 classes viz., Business, World, Sports, and Sci/Tech. The average size of each record is 39 words.
\item \textbf{BBC Sports} :
\cite{greene06icml} dataset consists of 737 sports articles each with an average size of 337 words. These articles are collected from the BBC Sport website and classified into 5 categories viz., Athletics, Cricket, Football, Rugby and Tennis. Using an 80:20 split, train data contains 590 records and test data contains 147 records.
\item \textbf{IMDB} :
\cite{maas-EtAl:2011:ACL-HLT2011} dataset is used for sentiment classification and contains 50000 movie reviews. It is a binary text analytics dataset with an average record size of 231 words. Each review is classified as either positive or negative. The standard train test split available for this dataset is 50\%.
\item \textbf{R8} :
\cite{R8:online} dataset is a subset of the Reuters 21578 dataset. Each record is divided into 8 categories viz., Earn, Acq, Trade, Ship, Grain, Crude, Interest, and Money-fx. The average size of each record is 64 words. Training data and test data consist of 5485 and 2189 documents respectively.
\end{itemize}

\begin{figure}
    \centering
    \includegraphics[scale=0.4]{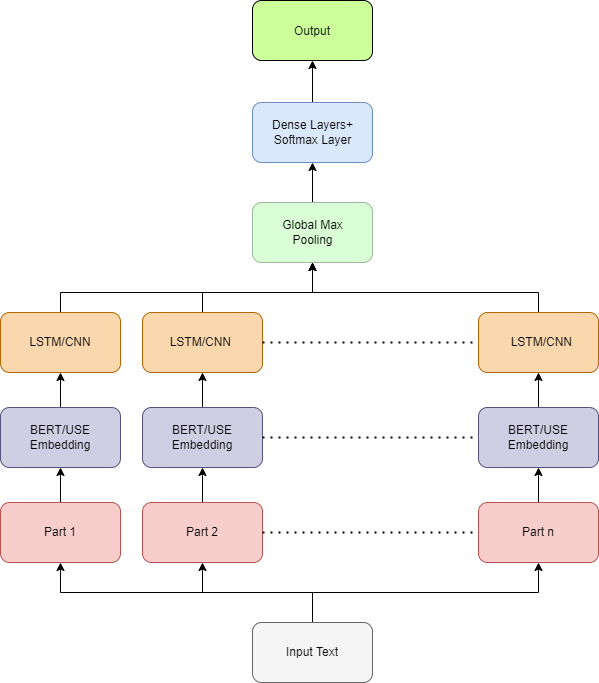}
    \caption{Block diagram of the proposed model architecture}
    \label{fig:block_diagram}
\end{figure}

\section{Experimental Setup}
The problem is formulated as a standard multi-class text classification problem. The input is pre-processed and the unwanted tokens and characters are removed. The clean text is then passed to the model with appropriate tokenization. The non-BERT based models use word tokenization whereas the BERT uses sub-word tokenization. For hierarchical models, the text is segmented into fixed-size chunks and the document is represented as a series of chunks. The total number of chunks in each document or sentence may vary depending upon the total words in it. The size of each chunk ranges from 20-50 words. The chunk size is chosen depending upon the average length of the dataset. The individual chunk is passed to base USE or BERT to get its sentence embedding which is further passed through a CNN or LSTM network. We have also experimented with other partitioning schemes like variable size chunks but the fixed-length chunks work optimally. 
\subsection{Preprocessing}
Following preprocessing steps are used on the input data:
\begin{itemize}
    \item Remove HTML tags: 
    HTML tags in the text data are removed as they often don't convey useful information.
    \item Normalization: 
    Complete text in the document is converted into lowercase.
    \item Remove accented characters: 
    In this step accented characters are converted into ASCII characters to prevent the NLP model from handling accented terms and others differently than their normal spellings.
    \item Expand Contractions: 
    Words that make use of apostrophes used for shortening text are expanded.
    \item Remove special characters: 
    Special characters are neither alphabets nor numerals, they are unreadable and hence need to be removed.
\end{itemize}

\section{Results}

\subsection{Architecture Details}
The USE, BERT, and HAN are the baseline architectures used in this work. We couple USE and BERT with CNN or LSTM to form a hierarchical model. The general setup of model is shown in Figure \ref{fig:block_diagram}. The hierarchical variations of the model are described below.
\begin{itemize}
    \item \textbf{USE + LSTM}: For this experiment, input data is converted into 512-dimensional vectors using the USE architecture. The pre-trained Universal sentence encoder from the TensorFlow library is used for generating high dimensional vectors which are then passed as input embeddings to an LSTM network. Initially, there are two Bi-LSTM layers with 256 and 128 units each, followed by a max-pooling layer. This is then linked to two series of dense, dropout, and batch normalization layers. Wherein the dense layers are of size 256 and 64 units with ‘Relu’ activation and the dropout rate is 0.4.\\
    
    \item \textbf{USE + CNN}: Similar to the previous architecture USE vectors are generated for the input data and this is connected as input embedding to the CNN network. This CNN network consists of 2 series of 512 units CNN layer with kernel size 1, max-pooling layer, and dropout layer of rate 0.5. This is then connected to a global max-pooling layer followed by 2 series of dense layers and dropout layers. These dense layers are of 1024 and 128 units each with ‘tanh’ activation and a dropout rate of 0.5 is employed. \\
    
    \item \textbf{BERT + LSTM}: In this architecture, we generate BERT embeddings for each dataset. The trained ‘bert-base-uncased’ model from the transformers library is used for generating embeddings. These embeddings are then connected to an LSTM network. This network firstly consists of 2 Bi-LSTM layers of 256 and 128 units respectively followed by a max-pooling layer. This is then connected to another dense layer with ‘Relu’ activation of 64 units. \\
    
    \item \textbf{BERT + CNN}: Similar to the previous architecture BERT embeddings are generated and passed as input to a CNN network. This CNN network consists of 2 CNN layers of 512 and 256 units each and a kernel size of 3, followed by a max-pooling layer and a dense layer of 64 units with ‘Relu’ activation.\\

\end{itemize}

\begin{table}
\begin{center}
\caption{Models and their accuracies on datasets}
\label{tab2}
 \begin{tabular*}{\textwidth}{c @{\extracolsep{\fill}} cccccc}
 \hline\hline\noalign{\smallskip}
\textbf{Model} & \textbf{20NG} & \textbf{BBC News} & \textbf{AG News} & \textbf{BBC Sports} & \textbf{IMDB} & \textbf{R8} \\
\hline\hline\noalign{\smallskip}
USE & 81.76 & 96.63 & 92.09 & 98.65 & 87.14 & 95.61\\
\hline\noalign{\smallskip}
BERT & \textbf{85.78} & 98.2 & \textbf{94.04} & 98.65 & 89.58 & \textbf{97.62}\\
\hline\noalign{\smallskip}
HAN & 85.01 & 97.75 & 92.11 & 96.24 & 88.94 & 94.47\\
\hline\noalign{\smallskip}
USE+LSTM & 81.81 & 98.2 & 92.25 & 98.65 & 88.89 & 95.75\\
\hline\noalign{\smallskip}
USE+CNN & 80.03 & 97.53 & 92.21 & 99.32 & 89.7 & 96.44\\
\hline\noalign{\smallskip}
BERT+LSTM & 85.57 & \textbf{98.43} & 94.01 & 99.32 & \textbf{93.63} & 95.89\\
\hline\noalign{\smallskip}
BERT+CNN & 83.79 & 98.2 & 92.4 & \textbf{100} & \textbf{93.63} & 96.35\\
\hline\noalign{\smallskip}
BigBird & 85.14 & 97.97 & 92.3 & 99.32 & \textbf{94.32} & \textbf{98.03}\\
\hline\noalign{\smallskip}
Longformer & \textbf{86.45} & \textbf{98.65} & 93.4 & \textbf{100} & 93.3 & 97.85\\
\hline\noalign{\smallskip}
 \end{tabular*}    
\end{center}

\end{table}

\section{Results}

In this work, we have obtained results on 6 different benchmark datasets. The primary algorithms used are USE, BERT, and HAN. The USE and HAN are non-transformer based algorithms and are expected to perform poorer than BERT-based models. The primary focus of this work is hierarchical approaches based on USE and BERT. The results of these algorithms are shown in Table \ref{tab2}. These results were explicitly re-computed on similar hyper-parameters and hardware environments to ensure a fair comparison. We compare USE + LSTM and USE + CNN with USE and HAN. We see that USE + CNN/LSTM performs better than HAN on all the datasets except for 20NG. Similarly, they either perform better than standalone USE or are on par with it. This shows that the hierarchical setup allows the model to capture better representations for classification.

The BERT model performs better than the USE and its variations on most of the datasets. However, BERT + CNN/LSTM performs on par with standalone BERT, unlike USE where there is a clear improvement in performance. Although the hierarchical setup does not provide an improvement over BERT it is still desirable because of its higher efficiency. Since we run the base BERT on individual chunks the self-attention is now restricted to words in the individual chunk. The complexity of self-attention is no longer quadratic in nature and this allows us to process longer-length sentences. Specifically, the BERT + LSTM model performs better in most cases. Overall we observe that the hierarchical setup is beneficial with both transformer and non-transformer approaches. 

The results for Longformer and BigBird are also added for comparison. These transformer architectures use specialized attention mechanisms to improve self-attention efficiency. Although Longformer does provide the best results on most of the datasets its attention mechanism hurts the parallelization of Transformers and requires specialized handling. Hence the vanilla BERT-based architectures are still relevant in this context.     

\section{Conclusion}
In this paper, we present four novel models of BERT + LSTM, USE + LSTM, USE + CNN, and BERT + CNN that were evaluated on 6 benchmark datasets. We observed that the BERT + LSTM model performs better than other extensions. The accuracies obtained from this are not only comparable to the baseline models for a few datasets like 20 newsgroups and AG news but it also outperforms the accuracies for BBC news and IMDB dataset. Similarly the USE + LSTM extension also either has improved results than baseline USE or has comparable results. Our results confirm that all four models can be efficiently used for the task of long document classification. We also provide a comparative analysis of deep learning algorithms like USE, BERT, HAN, Longformer, and BigBird for long document classification on these benchmark datasets.

\section*{Acknowledgements}
This research was conducted under the guidance of L3Cube, Pune. We would like to express our gratitude towards our mentors at L3Cube for their continuous support and encouragement.

\printbibliography









\end{document}